\documentclass{article}
\usepackage{ijcai25}

\usepackage{times}
\usepackage{soul}
\usepackage{url}
\usepackage[hidelinks]{hyperref}
\usepackage[utf8]{inputenc}
\usepackage[small]{caption}
\usepackage{graphicx}
\usepackage{amsmath}
\usepackage{amsthm}
\usepackage{booktabs}
\usepackage{algorithm}
\usepackage{algorithmic}
\usepackage[switch]{lineno}
\usepackage{multirow}
\usepackage{multicol}
\usepackage{xcolor}

\newcommand{\blue}[1]{\textcolor{blue}{#1}} 
\newcommand{\red}[1]{\textcolor{red}{#1}}   

\title{Improving Interpretability and Accuracy in Neuro-Symbolic Rule Extraction Using Class-Specific Sparse Filters}

\author{
Parth Padalkar
\and
Jaeseong Lee\and
Shiyi Wei\And
Gopal Gupta\\
\affiliations
The University of Texas at Dallas\\
\emails
\{parth.padalkar, jaeseong.lee, swei, gupta\}@utdallas.edu
}

\begin{document}

\maketitle

\begin{abstract}
There has been significant focus on creating neuro-symbolic models for interpretable image classification using Convolutional Neural Networks (CNNs). These methods aim to replace the CNN with a neuro-symbolic model consisting of the CNN, which is used as a feature extractor, and an interpretable rule-set extracted from the CNN itself. While these approaches provide interpretability through the extracted rule-set, they often compromise accuracy compared to the original CNN model.
In this paper, we identify the root cause of this accuracy loss as the post-training binarization of filter activations to extract the rule-set. To address this, we propose a novel sparsity loss function that enables class-specific filter binarization during CNN training, thus minimizing information loss when extracting the rule-set. We evaluate several training strategies with our novel sparsity loss, analyzing their effectiveness and providing guidance on their appropriate use. Notably, we set a new benchmark, achieving a $\textbf{9\%}$ improvement in accuracy and a $\textbf{53\%}$ reduction in rule-set size on average, compared to the previous SOTA, while coming within $\textbf{3\%}$ of the original CNN's accuracy. This highlights the significant potential of interpretable neuro-symbolic models as viable alternatives to black-box CNNs. 

\end{abstract}

\section{Introduction}
Interpretability in deep neural models has gained a lot of interest in recent years, e.g., \cite{interpretable_survey1,interpretable_survey2,interpretable_survey_3}. This is well placed, as some applications such as autonomous vehicles \cite{kanagaraj2021}, disease diagnosis \cite{Sun2016ComputerAL}, and natural disaster prevention \cite{Ko2012disaster} are very sensitive areas where a wrong prediction could be the difference between life and death. The above tasks rely heavily on good image classification models such as Convolutional Neural Networks (CNNs) \cite{cnn} which are not interpretable. Hence these applications could benefit greatly by using models that balance interpretability with decent accuracy.

Specifically, in the context of image classification using CNNs, there has been a lot of effort towards improving their interpretability by extracting a rule-set from the convolution layers, that explains the underlying decision-making logic of the model \cite{eric,eric2,nesyfold,nesyfold-g,nesybicor}. The rule-set along with the CNN upto the last convolutional layer forms the \textit{NeSy} model, where the final classification is done by the rule-set. This neuro-symbolic approach, where the neural component (CNN) and the symbolic component (rule-set) are used in conjunction for the classification task, has shown promise in areas such as interpretable Covid-19 and pleural effusion detection from chest X-ray images \cite{eric3}.

\begin{figure}[t]
    \centering
    \includegraphics[width = \linewidth, height = 7cm]{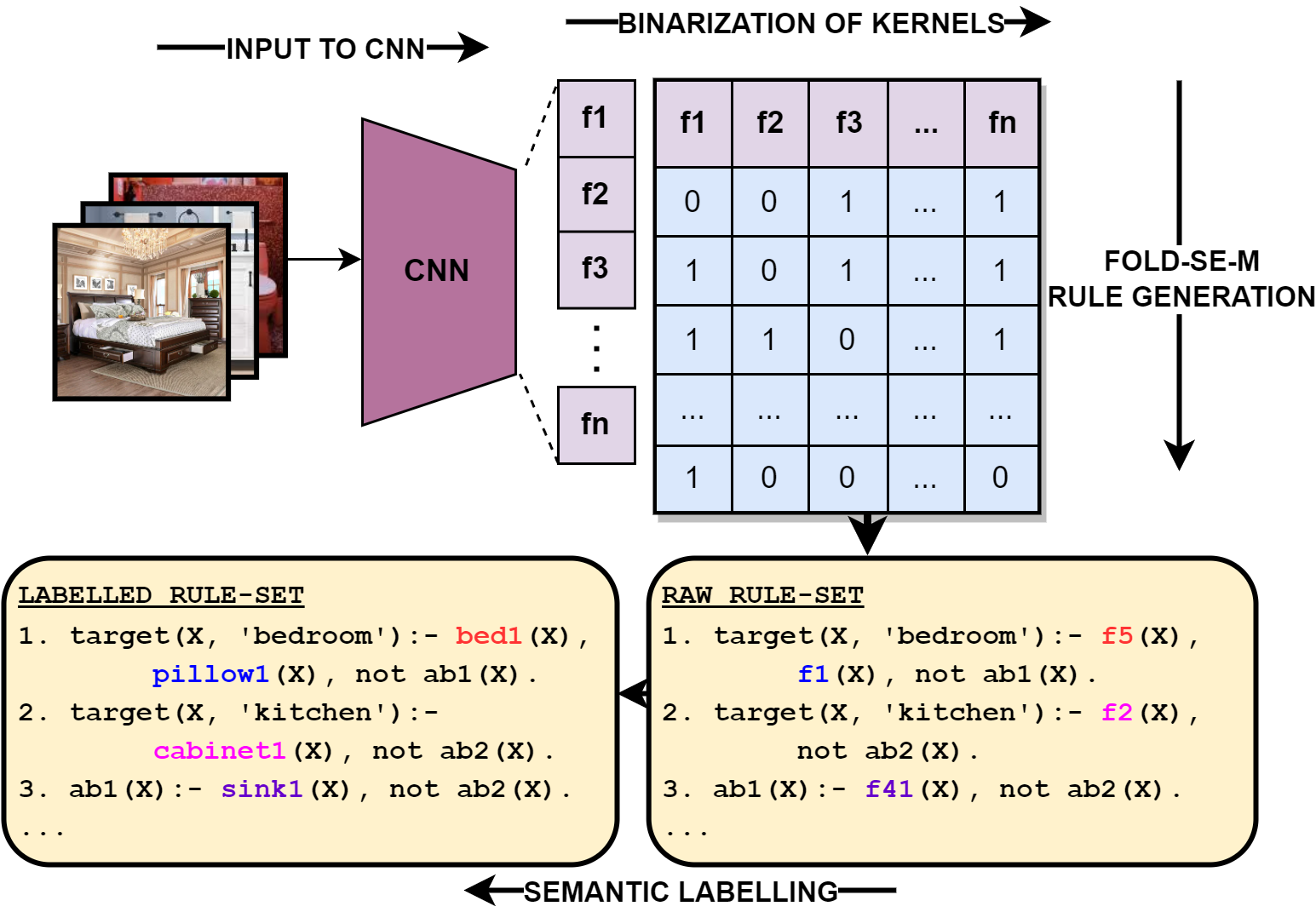}
    \caption{The NeSyFOLD Framework}
    \label{fig_nesyfold}
\end{figure}
The core of these neuro-symbolic methods lies in the binarization of the last-layer filter outputs using a threshold calculated post-training. This process converts each training image into a binarized feature vector derived from the final convolutional layer of the CNN. Subsequently, a decision-tree algorithm or rule-based machine learning algorithm is employed to generate symbolic rules from these binarized vectors. The classification is then performed by extracting the outputs of the last-layer filters, binarizing each filter's output using the calculated threshold, and applying the rule-set for the final classification. The binarized filter output represents the corresponding atom's/predicate's truth value in the rule-set. The final classification is done based on the rule that fires. Figure \ref{fig_nesyfold} illustrates the current state-of-the-art (SOTA), NeSyFOLD framework \cite{nesyfold}, for generating a symbolic rule-set from a CNN in the form of a logic program. Notice how all the filter outputs (f1, f2, ..., fn) are binarized for each image, creating a binarization table. This table along with the target class for each row, is fed to the FOLD-SE-M \cite{foldsem} rule-based machine algorithm that generates a rule-set in the form of a logic program. The filters appear as predicates in each rule's body and their binary value $(0/1)$ determines their respective predicate's truth value.

The filter outputs in a CNN are continuous and optimized during training to support the fully connected layers for final classification. However, in the aforementioned neurosymbolic methods the fully connected layers are essentially replaced with a rule-set for classification. Hence, these continuous outputs must be binarized or converted into a symbolic-friendly form $(0/1)$ to suit the requirements of the rule-extraction algorithm. This post-training binarization leads to information loss, as the filters were not originally optimized for such binarization.

One approach to mitigate this information loss is by training the model such that there are very few filters inside the CNN that are responsible for learning the representation of each class. This is known as ``\textit{learning class-specific sparse filters}" and is typically done using a loss function designed to induce sparsity \cite{EBP,icnn,iccnn}. 
While this approach reduces the number of filter outputs that require binarization post-training, it only partially addresses the issue, as the fundamental challenge of transitioning from continuous to symbolic binary representations still persists and hence there remains a significant gap between the accuracy of the original CNN model and the NeSy model. 

To address this limitation we propose a novel sparsity loss function which enforces a pre-selected subset of class-specific filter outputs, to converge towards values close to $1$ while pushing other filter outputs close to $0$. 
Hence, post-training, the pseudo-binarized filter outputs for each image can be directly rounded to create feature vectors, which are more optimal for post-training rule-extraction. We show through our experiments that using this loss function can lead to significant gains in the accuracy of the NeSy model ($9\%$) as compared to the previous SOTA (NeSyFOLD-EBP) while at the same time drastically reducing the size of the rule-set generated by $53\%$, thus improving interpretability. Thus, the accuracy gap between the trained CNN and the NeSy model is narrowed to an average of just $3\%$.
To summarize, our novel contributions are as follows:
\begin{enumerate}
    \item We introduce a novel sparsity loss function that can be used during training to learn sparse filters with pseudo-binarized outputs.
    \item We present a comprehensive analysis of $5$ training strategies using the sparsity loss and discuss their merits and pitfalls.
\end{enumerate}

\section{Background}
The filters in the CNN are matrices that capture patterns in the images. It has been shown that filters in the later layers of a CNN capture high-level concepts such as objects or object parts \cite{object_detectors_emerge}. Hence, this line of research has emerged, wherein the final decision-making rules are represented in terms of these filter outputs that learn high-level concept(s) in the images. 

\textit{NeSyFOLD} \cite{nesyfold} has emerged as a leading framework, extracting a logic program (specifically, stratified answer set program \cite{Baral}) from the CNN. Figure \ref{fig_nesyfold} illustrates the rule-extration pipeline of NeSyFOLD. The binarized outputs of the last convolutional layer, stored in the binarization table, serve as input to the FOLD-SE-M algorithm \cite{foldsem}. FOLD-SE-M then generates the raw rule-set where the truth value of that predicate is determined by the corresponding filter's binarized output i.e. $0$ or $1$ (Fig. \ref{fig_nesyfold}, bottom-right). Then the filters are matched to the concepts that they have learnt from the images using a semantic labelling algorithm that uses the images' semantic segmentation masks (which are masks of the images with every pixel labelled with the concept that it belongs to) to determine what concept(s) the filters are most activated by in the images. The predicates in the raw rule-set are then labelled as those concept(s). Then, during test-time, the images are passed through the CNN and their respective filter outputs are binarized using the threshold computed after training. Based on the binary filter activation values, the truth value of the predicates in the rule-set is determined and thus the classification is made based on the rule that evaluates to true.



FOLD-SE-M \cite{foldsem} is a Rule-Based Machine Learning (RBML) algorithm that generates rules from tabular data, encoding them as stratified answer set programs. This means that there are no cycles through negation in the rules. It uses special \texttt{abx} predicates to represent exceptions, where \texttt{x} is a unique identifier. The algorithm incrementally generates literals for default rules to cover positive examples while avoiding negative ones. It recursively learns exceptions by swapping positive and negative examples. FOLD-SE-M includes two hyperparameters: \textit{ratio}, which limits the ratio of false positives to true positives covered by a predicate, and \textit{tail}, which sets a threshold for the minimum number of examples covered by a rule. Compared to decision-tree classifiers, FOLD-SE-M has been shown to produce fewer, more interpretable rules with higher accuracy \cite{foldsem}.

The NeSyFOLD framework when used with a CNN trained using a class-specific sparse filter learning technique such as \textit{Elite BackProp (EBP)} \cite{EBP} achieves state-of-the-art results \cite{nesyfold}.
EBP is designed to associate each class with a small number of highly responsive ``Elite" filters. This is achieved by employing a loss function (along with the cross-entropy loss) that penalizes filters with a lower probability of activation for any class, while reinforcing those with higher activation probabilities during training. As a result, only a few filters learn the representations for each class. The number of elite filters assigned to each class is controlled by the hyperparameter \( K \).
This reduces the number of filters that need to be binarized post-training which somewhat reduces the information loss due to fewer filter output binarizations required. However, there is no focus on optimizing the filter outputs to a more symbolic-friendly, binary form during training so there is still significant information loss when a rule-set is extracted. 


\section{Methodology}
\subsection{Calculating the Sparsity Loss}
We now introduce our novel sparsity loss for learning class-specific filters and effective binarization of those filter outputs to address the limitation of EBP discussed above.

\subsubsection{Computing the Filter Probability Matrix ($P$)}

The \( P \) matrix stores the probability of activation for each filter across all classes.  It is a 2D matrix of shape \( (C, F) \), where $C$ is the total number of classes and $F$ is the total number of filters in the last convolutional layer.
We indicate $2$ different methods to compute the $P$ matrix.

\noindent\underline{\textit{Method 1 (Using class-specific activation frequency):}}
\begin{enumerate}
    \item Extract the feature maps generated by each filter from the last convolutional layer for all training images.
    \item For each image, compute the normalized feature map output for each filter:
    \[
    \text{Norm}_{i,j} = \frac{\sum_{h,w} |f_{i,j,h,w}|}{H \cdot W}, \quad i \in [1, N], \, j \in [1, F]
    \]
    where \( H \) and \( W \) are the spatial dimensions of the feature map $f$, \( N \) is the number of training images, and \( F \) is the number of filters.
    \item Accumulate these norms into a class-filter matrix \( D \) of shape $(C, F)$ by summing over all images of the same class such that each row represents a class and each column represents the cumulative norm value for each filter.
    \item For each class, identify the top \( K \) filters with the highest cumulative activations, where \( K \) is a hyperparameter. 
    \item Calculate probability \( P[i][j] \) for each filter \( j \) in class \( i \):
    \begin{align}\label{eq_P}
        P[i][j] = 
            \begin{cases} 
            1, & \text{if } j \in \text{Top-}K \text{ filters for } i \\
            0, & \text{otherwise}
            \end{cases}    
    \end{align}

\end{enumerate}

\noindent\underline{\textit{Method 2 (Random Initialization):}}
Compute the $P$ matrix by randomly initializing $K$ filters for each class as $1$ and every other filter as $0$.

\subsubsection{Threshold Tensor Calculation}
The threshold tensor is computed to determine the activation thresholds for each filter. The steps are:
\begin{enumerate}
    \item For each training image, calculate the \( L_2 \)-norm of the filter feature maps from the last convolutional layer:
    \[
    \text{Norm}_{i,j} = \|f_{i,j}\|_2, \quad i \in [1, N], \, j \in [1, F]
    \]
    \item Compute the mean (\( \mu_j \)) and standard deviation (\( \sigma_j \)) of these norms for each filter across all images.
    \item Calculate the threshold for each filter:
    \begin{align}\label{eq_threshold}
        \text{Threshold}_j = h_1 \cdot \mu_j + h_2 \cdot \sigma_j    
    \end{align}

    where \( h_1 \) and \( h_2 \) are hyperparameters.
\end{enumerate}

\subsubsection{Sparsity Loss Computation}

During training, the sparsity loss is calculated as follows:
\begin{enumerate}
    \item Compute the \( L_2 \)-norms of the filter feature maps for every input image $n$:
    \begin{align}\label{eq_norm}
        \text{Norm}_{n,j} = \|f_{n,j}\|_2, \quad n \in [1, N], \, j \in [1, F]    
    \end{align}
    
    \item Subtract the precomputed threshold from each filter norm:
    \begin{align}\label{eq_adjusted_norm}
        \text{Adjusted}_{n,j} = \text{Norm}_{n,j} - \text{Threshold}_j  
    \end{align}
    
    \item Apply the sigmoid function to the adjusted norms so that the outputs are in range $[0, 1]$:
    \begin{align}\label{eq_sigmoid}
        \text{Sigmoid}_{n,j} = \sigma(\text{Adjusted}_{n,j}) = \frac{1}{1 + e^{-\text{Adjusted}_{n,j}}}    
    \end{align}
    
    \item Retrieve the filter activation probabilities for the target class of each image using the \( P \) matrix:
    \[
    \text{ClassProbabilities}_{n,j} = P[\text{class of }n][j]
    \]
    \item Define the target activations for the filters for each image:
    \begin{align}\label{eq_filter_targets}
        \text{Target}_{n,j} = 
        \begin{cases} 
        1, & \text{if } \text{ClassProbabilities}_{n,j} = 1 \\
        0, & \text{otherwise}
        \end{cases}
    \end{align}
    
    \item Compute the Binary Cross-Entropy (BCE) loss between the predicted sigmoid activations and the target activations:
    \begin{align}
        \mathcal{L}_{\text{sparsity}} &= - \frac{1}{N \cdot F} 
        \sum_{n=1}^{N} \sum_{j=1}^{F} 
        \bigg[
        \text{Target}_{n,j} \cdot \log(\text{Sigmoid}_{n,j}) \notag \\
        &\quad + (1 - \text{Target}_{n,j}) \cdot \log(1 - \text{Sigmoid}_{n,j})
        \bigg]
    \end{align}

\end{enumerate}

\subsubsection{Total Loss}

The sparsity loss is combined with the cross-entropy loss to form the total Loss $\mathcal{L}$:
\begin{align}\label{eq_final_loss_fn}
    \mathcal{L} = \alpha \cdot \mathcal{L}_{\text{cross-entropy}} + \beta \cdot \mathcal{L}_{\text{sparsity}}    
\end{align}

where \( \alpha \) and \( \beta \) are hyperparameters controlling the trade-off between classification accuracy and filter sparsity.

The most critical step in calculating the sparsity loss is assigning each filter a target value of either \(1\) or \(0\) (Equation \eqref{eq_filter_targets}). This assignment frames the problem as a binary classification task, where each filter is categorized as either ``active" or ``inactive." The Binary Cross-Entropy (BCE) loss is then applied to optimize this classification which is designed for such tasks. Thus, as the training progresses, the filter outputs gradually converge toward binary values (0 or 1). This ensures minimal information loss when the outputs are finally rounded to binarize them for rule-extraction.

\subsection{Extracting the Rule-Set and Inference}
Once the training is complete, the NeSyFOLD pipeline is employed wherein each image in the train set is passed through the CNN and the last layer filter feature maps are obtained. Then Equations \eqref{eq_norm} to \eqref{eq_sigmoid} are applied in order to obtain the sigmoid values which are rounded to the nearest integer (i.e. $0$ or $1$). Thus, each image is converted to a binarized vector and the FOLD-SE-M algorithm is used on the full binarized train set to obtain the rule-set.

At test-time, the input images are passed through the CNN to obtain the outputs of the last convolutional layer filters. The rounded sigmoid values are then computed as described previously. These values are used as truth values for the predicates in the rule-set. The final classification is made by employing the FOLD-SE-M toolkit's internal interpreter to determine which rule is activated based on these truth values.

\section{Experiments and Results}
We conducted experiments to address the following research questions:\\
\noindent\textbf{Q1:} How does altering various steps in the sparsity loss computation affect the performance of the NeSy model?\\
\noindent\textbf{Q2:} What is the maximum performance gain that can be achieved w.r.t. accuracy and rule-set size compared to NeSyFOLD-EBP, using the sparsity loss?\\
\noindent\textbf{Q3:} How well does this approach scale as the number of classes increases?\\
\medskip\noindent\textbf{Q4:} What effect does the sparsity loss have on the representations learned by the CNN filters?\\
\noindent\textbf{[Q1, Q2, Q3] Training Strategies (TS), Performance and Scalability:}

We evaluate various training strategies, each by varying a key step in the computation of the sparsity loss. First, we explain the setup of our experiments:

\noindent\textbf{Setup: } We evaluate performance using three key metrics: (1) the accuracy of the NeSy model (comprising the CNN and the extracted rule-set), (2) the fidelity of the NeSy model with respect to the original CNN, and (3) the total number of predicates in the rule-set (referred to as the rule-set size). Fidelity is defined as the proportion of predictions made by the NeSy model that match those of the original CNN, calculated by dividing the number of matching predictions by the total number of images in the test set. A smaller rule-set size improves interpretability \cite{rulesetinterpretability}, hence we use rule-set size as a metric of interpretability.

\noindent\underline{\textit{Datasets: }}We evaluate our approach on the same datasets as NeSyFOLD-EBP (NeSyFOLD with Elite BackProp (EBP)) \cite{nesyfold}, ensuring a fair comparison. We used the \textit{Places} \cite{places} dataset which contains images from various indoor and outdoor ``scene" classes such as ``bathroom", ``bedroom", ``desert road", ``forest road" etc. We created multiple subsets from this dataset of varying number of classes. \textbf{P2} includes images from the \textit{bathroom} and \textit{bedroom} classes, \textbf{P3.1} is formed by adding the \textit{kitchen} class  images to P2. \textbf{P5} is created by adding \textit{dining room} and \textit{living room} images to P3.1, and \textbf{P10} further includes \textit{home office, office, waiting room, conference room,} and \textit{hotel room} images in addition to all classes in P5. Additionally, \textbf{P3.2} comprises \textit{desert road, forest road}, and \textit{street} images, while \textbf{P3.3} contains \textit{desert road, driveway}, and \textit{highway} images. Each class has $5k$ images of which we made a $4k/1k$ train-test split for each class and we used the given validation set as it is. 
We also used the \emph{German Traffic Sign Recognition Benchmark} (GTSRB) \cite{gtsrb} dataset which consists of images of various traffic signposts. This dataset has $43$ classes of signposts. We used the given test set of $12.6k$ images as it is and did an $80:20$ train-validation split which gave roughly $21k$ images for the train set and $5k$ for the validation set.

\noindent\underline{\textit{Hyperparameters: }}We employed a VGG16 CNN pretrained on \textit{ImageNet} \cite{deng2009imagenet}, training for $100$ epochs with batch size $32$. The Adam \cite{adamoptim} optimizer was used, accompanied by class weights to address data imbalance. $L2$ Regularization of $0.005$ spanning all layers, and a learning rate of $5 \times 10^{-6}$ was adopted. A decay factor of $0.5$ with a $10$-epoch patience was implemented. Images were resized to $224 \times 224$, and hyperparameters $h1$ and $h2$ (eq. \eqref{eq_threshold}) for calculating threshold for binarization of kernels, were set at $0.6$ and $0.7$ respectively. The $\alpha$ and $\beta$ used in the final loss calculation (Eq. \eqref{eq_final_loss_fn}) were set to $1$ and $5$ respectively. The $K$ value to find \textit{Top-K} filters per class as described in Method 1, step $4$ of computing the filter probability matrix ($P$) is set to $5$ for P2, P3.1, P3.2 and P3.3 and $20$ for P10 and GT43 as they have a higher number of classes.
The $ratio$ and $tail$ hyperparameter values for rule extraction using FOLD-SE-M were set to $0.8$ and $5 \times 10^{-3}$ respectively.

Next, we discuss the various training strategies using this sparsity loss function

\medskip\noindent\textbf{TS 1:} This strategy involves training the CNN for $50$ epochs with the sparsity loss term set to $0$. After the $50^{th}$ epoch we compute the $P$ matrix using Method 1, \textit{Top-K} filters per class and the thresholds for all the filters. Then, we train for $50$ more epochs with the sparsity loss term in effect. The intuition behind this approach is that training the CNN without sparsity constraints in the initial phase should allows the filters to naturally specialize and differentiate themselves. This specialization should result in a more informed selection of the \textit{Top-K} filters, whose targets are set to $1$ during the subsequent training phase. This two-step process aims to improve both accuracy and the quality of the learned representations.

\medskip\noindent\textbf{TS 2:} Here we compute the filter thresholds, $P$ matrix using Method $1$, and \textit{Top-K} from the initial ImageNet pretrained weights of the CNN. The sparsity loss is employed from the start along with the cross-entropy loss. This is done to understand how enforcing sparsity constraints from the beginning of training affects the accuracy and rule-set size.

\medskip\noindent\textbf{TS 3:} To understand how the choice of the \textit{Top-K} filters affects the performance, we initialize $P$ using the random initialization method (Method 2). So, \textit{Top-K} filters for each class are chosen as these randomly assigned filters and the sparsity loss is employed from the start along with the cross-entropy loss.

\medskip\noindent\textbf{TS 4:} In this strategy, we use the same configuration as \textbf{TS 3}, but the cross-entropy loss term is set to 0. Consequently, only the sparsity loss is optimized throughout the training. This approach is designed to assess whether learning class-specific filters alone provides sufficient information for the FOLD-SE-M algorithm to partition the feature space and generate effective rules for each class.

\medskip\noindent\textbf{TS 5:} Similar to \textbf{TS 3}, this strategy keeps the same configuration but skips the computation of thresholds. Instead, the sigmoid function is directly applied to the computed norms in Eq. \eqref{eq_norm} without subtracting the thresholds. This approach helps in understanding the impact of threshold computation on the training process and the overall performance.

\begin{table}[t]
\centering
\fontsize{9.5}{10}\selectfont 
\setlength{\tabcolsep}{2.5pt} 
\begin{tabular}{@{}lrrrrrrrr@{}}
\toprule
     & P2              & P3.1            & P3.2            & P3.3            & P5              & P10             & GT43            & \textbf{MS}              \\ \midrule
NE   & \blue{92}, \red{12} & \blue{86}, \red{16} & \blue{91}, \red{07}  & \blue{78}, \red{23} & \blue{67}, \red{30} & \blue{44}, \red{65} & \blue{85}, \red{99} & \blue{78}, \red{36} \\
TS1 & \blue{94}, \red{10} & \blue{87}, \red{17} & \blue{93}, \red{13} & \blue{82}, \red{24} & \blue{61}, \red{19} & \blue{44}, \red{50} & \blue{80}, \red{76} & \blue{77}, \red{30} \\
TS2 & \blue{96}, \textbf{\red{05}} & \blue{92}, \red{11} & \blue{94}, \textbf{\red{07}} & \blue{85}, \red{18} & \textbf{\blue{78}}, \red{22} & \textbf{\blue{63}}, \red{35} & \blue{91}, \red{67} & \blue{86}, \red{24}\\

TS3 & \textbf{\blue{98}}, \textbf{\red{05}} & \textbf{\blue{95}}, \red{07} & \textbf{\blue{97}}, \textbf{\red{07}} & \textbf{\blue{89}}, \red{18} & \blue{75}, \red{08} & \blue{61}, \red{29} & \textbf{\blue{95}}, \red{43} & \textbf{\blue{87}}, \red{17} \\ 
TS4 & \blue{94}, \textbf{\red{05}} & \blue{86}, \textbf{\red{05}} & \blue{93}, \textbf{\red{07}} & \blue{82}, \textbf{\red{11}} & \blue{62}, \textbf{\red{07}} & \blue{48}, \red{16} & \blue{90}, \red{44} & \blue{79}, \red{14} \\
TS5  & \blue{94}, \red{08} & \blue{88}, \red{16} & \blue{89}, \red{11} & \blue{74}, \red{17} & \blue{56}, \red{25} & \blue{08}, \textbf{\red{01}} & \blue{17}, \textbf{\red{06}} & \blue{61}, \textbf{\red{12}} \\ \bottomrule
\end{tabular}
\caption{Accuracy (blue) and the rule-set size (red) of the NeSy model generated by each Training Strategy (TSx). NE is NeSyFOLD-EBP. The headers are various datasets and \textbf{MS} shows the Mean Statistics across all datasets.}
\label{tb_acc_size}
\end{table}
\begin{table}[t]
\centering
\fontsize{9.5}{10}\selectfont 
\setlength{\tabcolsep}{2.5pt} 
\begin{tabular}{@{}lrrrrrrrr@{}}
\toprule
     & P2              & P3.1            & P3.2            & P3.3            & P5              & P10             & GT43            & \textbf{MS}              \\ \midrule
NE   & \blue{97}, \red{93} & \blue{94}, \red{87} & \blue{96}, \red{92}  & \blue{89}, \red{82} & \blue{85}, \red{70} & \blue{70}, \red{49} & \blue{98}, \red{85} & \blue{90}, \red{80} \\ 
TS1  & \blue{97}, \red{95} & \blue{94}, \red{89} & \blue{96}, \red{94} & \blue{88}, \red{85} & \blue{85}, \red{63} & \blue{70}, \red{48} & \blue{98}, \red{80} & \blue{90}, \red{79} \\
TS2 & \blue{97}, \red{97} & \blue{94}, \red{94} & \blue{96}, \red{96} & \blue{88}, \textbf{\red{91}} & \blue{84}, \textbf{\red{78}} & \blue{69}, \textbf{\red{66}} & \blue{97}, \red{90} & \blue{89}, \textbf{\red{87}} \\

TS3  & \blue{97}, \textbf{\red{98}} & \blue{94}, \textbf{\red{95}} & \blue{96}, \textbf{\red{97}} & \blue{88}, \red{89} & \blue{85}, \red{75} & \blue{69}, \red{63} & \blue{98}, \textbf{\red{95}} & \blue{90}, \textbf{\red{87}} \\ 
TS4  & \blue{50}, \red{49} & \blue{33}, \red{30} & \blue{33}, \red{32}  & \blue{33}, \red{32} & \blue{20}, \red{15} & \blue{10}, \red{07} & \blue{03}, \red{03} & \blue{26}, \red{24} \\ 
TS5  & \blue{96}, \red{94} & \blue{92}, \red{89} & \blue{96}, \red{90}  & \blue{87}, \red{75} & \blue{80}, \red{58} & \blue{21}, \red{00} & \blue{94}, \red{17} & \blue{81}, \red{60} \\ \bottomrule
\end{tabular}
\caption{The CNN model's Accuracy (blue) and the Fidelity (red) of the NeSy mdoel generated for each Training Strategy (TSx). NE is NeSyFOLD with EBP. The headers are various datasets and \textbf{MS} shows the Mean Statistics across all datasets.}
\label{tb_CNNacc_fid}
\end{table}
\begin{figure*}[t]
    \centering
    \includegraphics[width = \linewidth, height = 8cm]{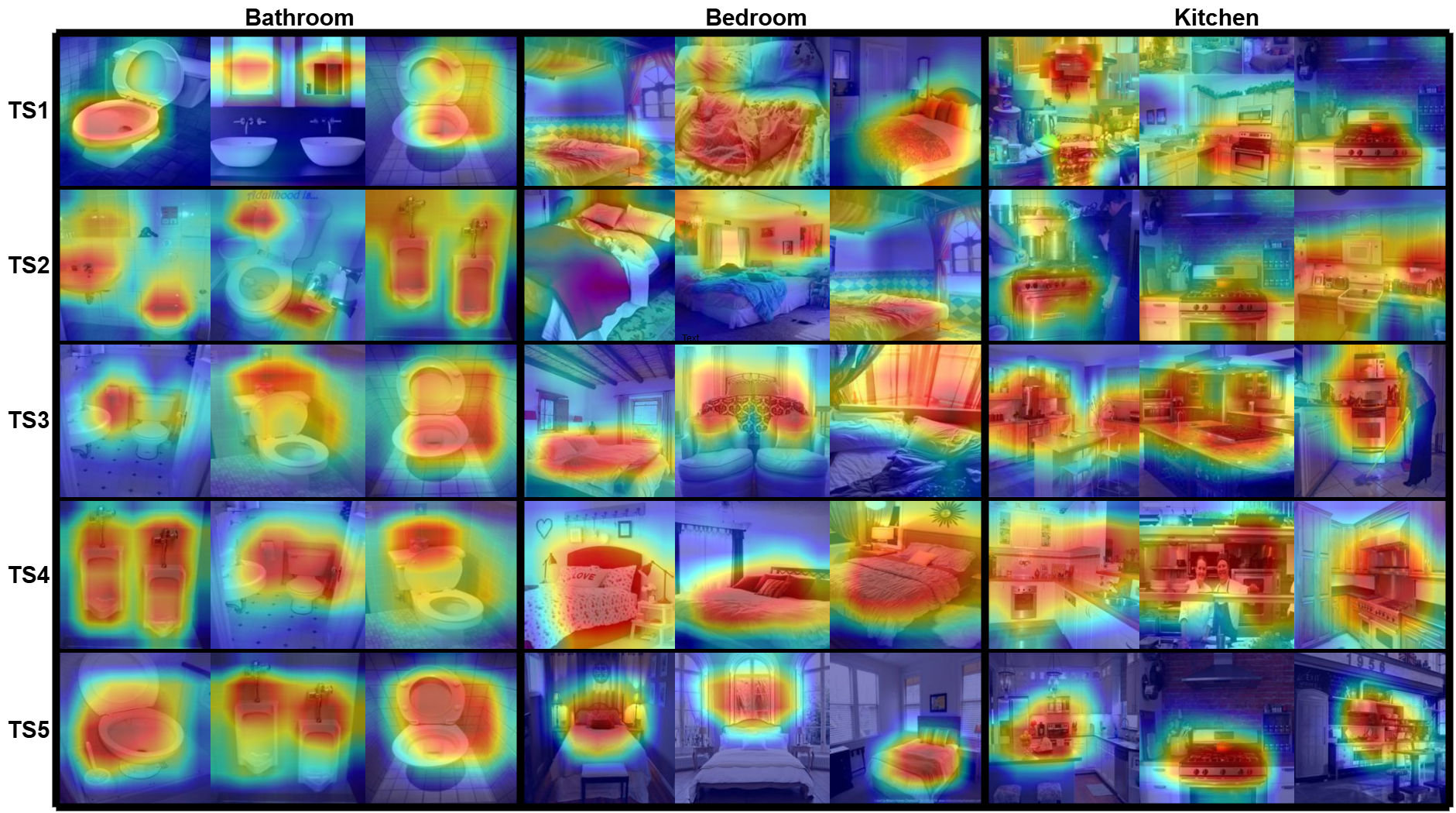}
    \caption{Activation maps of the top filters for the training strategies \textbf{TS1 - TS5} for P3.1 dataset. Each row represents the $top-3$ images for the top filter, per class, per training strategy. }
    \label{fig_heatmaps}
\end{figure*}
\noindent\textbf{[Results: Q1, Q2] Performance Comparison among various Training Strategies and Maximum Gain: }
We present the accuracy of the NeSy model (\blue{blue}) and the rule-set size (\red{red}) for all training strategies and NeSyFOLD-EBP (NE) in Table \ref{tb_acc_size}. Table \ref{tb_CNNacc_fid} reports the original CNN model's accuracy (\blue{blue}) alongside the NeSy model's fidelity with respect to the CNN (\red{red}). Accuracy and fidelity are average percentages, while rule-set size represents the average rule-set size, all computed over $5$ runs per dataset and rounded to the nearest integer. \textbf{MS} shows the average over all datasets for each strategy. When discussing the performance of each strategy, we omit specifying that it pertains to the NeSy model generated by that strategy, unless explicitly referring to the original CNN model.

Table \ref{tb_acc_size} shows that TS2, TS3, and TS4 outperforms NE in both accuracy and rule-set size. TS1 achieves accuracy comparable to NE while generating a smaller rule-set, on average. TS5 performs the worst w.r.t. accuracy but generates the smallest rule-sets. Ideally, the accuracy-to-rule-set-size ratio should be maximized, favoring high accuracy with smaller rule-sets for better interpretability.

TS3 stands out as the best overall, achieving $\textbf{9\%}$ higher accuracy and generating a rule-set that is $\textbf{53\%}$ smaller than NE on average. TS3 does even better than the trained CNN on P2, P3.1, P3.2 and P3.3 w.r.t. accuracy.

In TS3, $K$ filters are randomly assigned a probability of $1$ per class, with the rest set to $0$ in the $P$ matrix at the start of training. Interestingly, this approach outperforms TS2, which achieves a $\textbf{8\%}$ higher accuracy and a $\textbf{33\%}$ reduction in rule-set size compared to NE. Recall that in our experiments the CNN was initialized with pretrained ImageNet weights, as is standard practice. 

In TS2, the \textit{Top-K} filters are selected at the start of the training based on their activation strength using feature map norms. However, the poorer performance compared to TS3 may stem from the pretrained ImageNet weights, which are obtained through conventional training with a cross-entropy loss. This process might not yield an optimal filter selection for the sparsity loss and hence the random selection of \textit{Top-K} filters in TS3 leads to better performance. 

This trend is even more pronounced in TS1, where the CNN is initially trained for $50$ epochs using only the cross-entropy loss. Afterward, the \textit{Top-K} filters and the $P$ matrix are computed, and the sparsity loss is activated alongside the cross-entropy loss for another $50$ epochs. However, since the initial filter selection is based on a CNN trained solely with cross-entropy loss, the chosen \textit{Top-K} filters might not be optimal for the sparsity loss, thus leading to a suboptimal performance.

In TS4, the CNN is trained solely with the sparsity loss, without the cross-entropy loss, using randomly initialized \textit{Top-K} filters. This leads to a remarkable observation: even without cross-entropy, the NeSy model achieves a $\textbf{1\%}$ accuracy gain and a $\textbf{61\%}$ reduction in rule-set size compared to NE. The absence of cross-entropy constraints allows the sparsity loss to better separate filters in the latent space, optimizing their utility as features in the binarization table. As a result, FOLD-SE-M can effectively generate rules that segregate classes using fewer filters per class, reducing the rule-set size while maintaining high accuracy.

In TS5, the thresholds are not calculated, and \textit{Top-K} filters are randomly selected. The sigmoid is applied directly to the norms of the filters without subtracting thresholds beforehand. As norms are always positive, the minimum sigmoid value becomes 0.5, limiting the representation power. This means that the non-\textit{Top-K} (both poorly and highly activating) filters can only be assigned a value of 0.5 at the minimum, making them harder to distinguish from the \textit{Top-K} filters. In contrast, subtracting the threshold enables some filter norms to become negative, allowing the sigmoid to push irrelevant filters closer to 0. This facilitates true binarization, where non-relevant filters are suppressed, and preferably the \textit{Top-K} filters remain closer to 1. The lack of threshold subtraction in TS5 compromises its performance especially as the number of classes increases.

Note that the accuracy of the original CNN model as denoted in Table \ref{tb_CNNacc_fid} (\blue{blue}) is similar for NE and TS1-TS3. TS4 doesn't employ the cross-entropy loss so the accuracy of the CNN is naturally low. In TS5 the accuracy drops as the number of classes increases. Also note that the gap between the original CNN model and the NeSy model in terms of accuracy is $12\%$ in NE but it drops to just $\textbf{3\%}$ and $\textbf{2\%}$ for TS3 and TS2 respectively, suggesting that the information loss caused by post-training binarization is greatly reduced.

The NeSy models generated in both TS2 and TS3 have the highest average fidelity which is $\textbf{7\%}$ higher than NE. TS4 shows the lowest fidelity because the CNN is not trained for classification. So, the NeSy model generated by TS4 is a stand alone-model that does not follow the original model at all but does give a higher accuracy than the baseline (NE) on average. TS5 shows poor fidelity because of the limited representational capability as discussed earlier.

\noindent\textbf{[Results: Q3] Scalability: } TS2, TS3, and TS4 perform better than NE for P10 and GT43 which are datasets with $10$ and $43$ classes respectively, showing better scalability. In fact, TS2 and TS3 show a $\textbf{19\%}$ and $\textbf{17\%}$ increase in accuracy and show a $\textbf{46\%}$ and $\textbf{55\%}$ decrease in rule-set size respectively for P10. Similarly for GT43, TS2 and TS3 show a $\textbf{6\%}$ and $\textbf{10\%}$ increase in accuracy and a $\textbf{32\%}$ and $\textbf{57\%}$ decrease in rule-set size.

\noindent\textbf{[Q4] Filter Representations:}\\
\noindent\textbf{Setup:}For the P3.1 dataset, which includes the \textit{bathroom, bedroom, and kitchen} classes, we analyzed the extracted rule-sets for TS1–TS5. For each class, under each training strategy, we identified the top filter by examining the rule-set and selecting the filter that appeared as the first non negated predicate in the first rule for each class. We then overlaid the feature maps of these filters onto the top three images that activated them the most. Figure \ref{fig_heatmaps} displays these overlays, showing three images per filter, per class, and per training strategy.
Each row corresponds to a training strategy, while each set of three columns represents the top three images that most activate the chosen filter for a specific class. For instance, filters $145$, $295$, and $38$ were selected for TS3 for the \textit{bathroom}, \textit{bedroom}, and \textit{kitchen} classes, respectively. The first three columns show the top three images activating filter $145$, the next three columns correspond to filter $295$, and the final three columns show the images activating filter $38$.

\noindent\textbf{[Result: Q4]: } The filters typically activate for concepts relevant to their respective classes. For instance, the \textit{bathroom} filter consistently activates for toilets and sinks, the \textit{bedroom} filter for beds, and the \textit{kitchen} filter for stoves and cabinets. Notably, TS4 filters also activate for these relevant concepts, even though the CNN in this strategy was not trained with cross-entropy loss for classification. This demonstrates that our sparsity loss alone is sufficient to guide the filters in learning meaningful concept representations.

\begin{figure}[h!]
    \centering
    \includegraphics[width = \linewidth, height = 11cm]{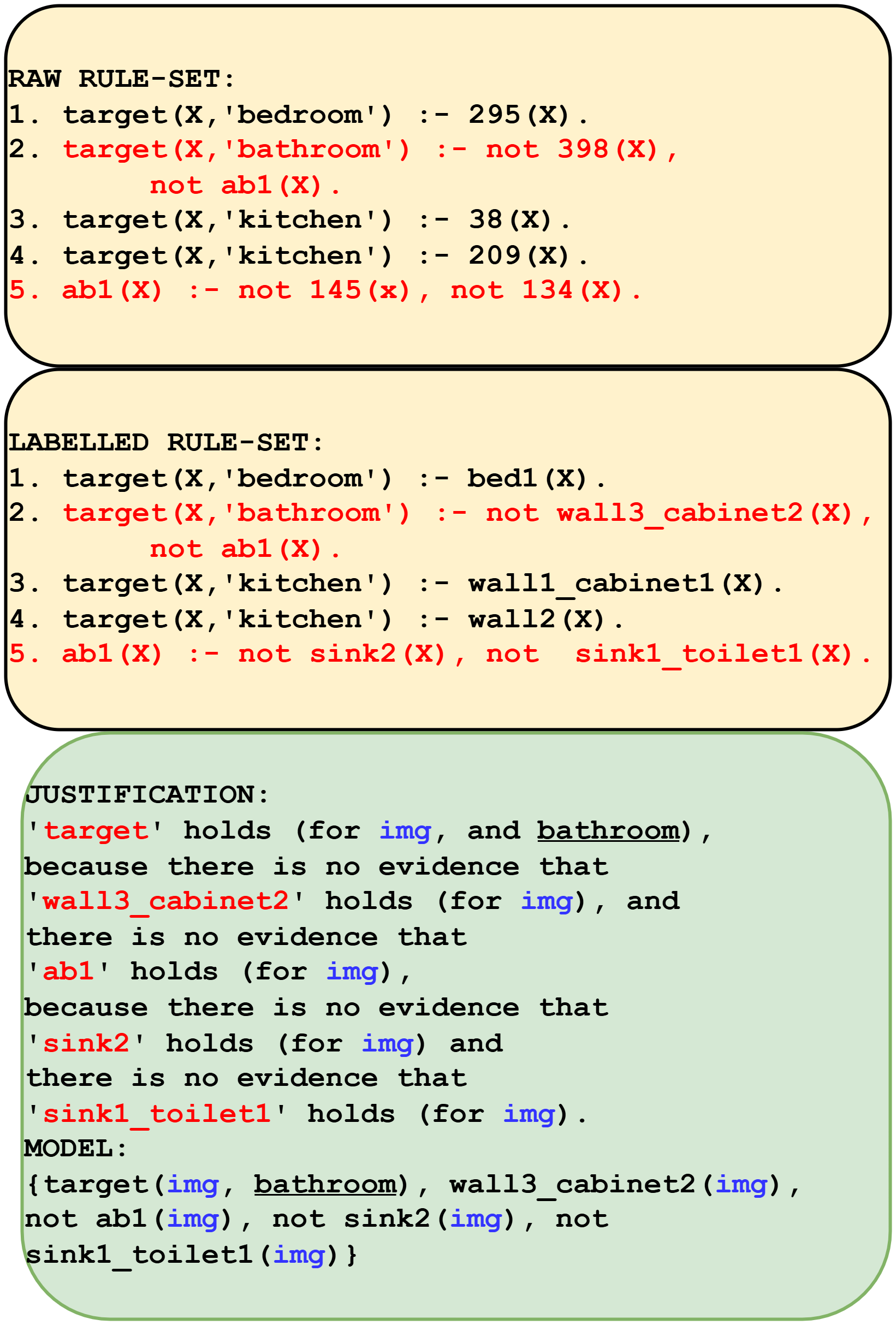}
    \caption{Raw rule-set produced via TS3 for P3.1 dataset (top). Labelled rule-set produced via the NeSyFOLD toolkit (middle) and justification provided by s(CASP) ASP engine for an image (bottom).}
    \label{fig_labelled_rule_set}
\end{figure}

Using such overlays, and provided semantic segmentation masks the NeSyFOLD toolkit allows to label each predicate in the rule-set with the concept(s) that it represents, thus creating a labelled rule-set that is highly interpretable. Figure \ref{fig_labelled_rule_set} (top) shows one such raw rule-set generated for the P3.1 dataset using TS3. Then via semantic labelling using the NeSyFOLD toolkit each predicate is mapped to the concept(s) its corresponding filter represents as shown in Figure \ref{fig_labelled_rule_set} (middle).  Recall that the rule-set is a stratified ASP program and thus an ASP interpreter such as s(CASP) \cite{scasp} can be used to obtain justification of an image represented as facts corresponding to the activated (1) binarized filters such as \{{\tt 145(img). 134(img).}\} etc. Then using the query, {\tt ?-target(img, X).} against a rule-set such as the one shown in Figure \ref{fig_labelled_rule_set} (top), a model (i.e. answer for {\tt X} or class of {\tt img}), as well as a justification, can be obtained (Figure \ref{fig_labelled_rule_set} (bottom)). Notice that the filter {\tt 398} is labelled as {\tt wall3\_cabinet2} signifying that this filter is activated by a particular type of wall-cabinet combination in images. Also, the numbers in the label of the predicate are just indicators that other predicates have also been labelled as similar concepts. This showcases the advantage of interpretable methods as every decision made by the model can be traced in a systematic manner. 

\section{Related Works}
Extracting knowledge from neural networks is a well studied topic \cite{ANDREWS1995373,tickle1998truth} but only recently extracting rules from CNNs by binarizing the filter outputs has gained popularity \cite{eric,eric2,nesyfold}. However, none of these approaches facilitate binarization of filter outputs like our method. Our approach addresses the performance gap between the NeSy model and the trained CNN.

There has been a lot of effort in sparsifying the weights of neural networks \cite{sparsity_survey}. However, our work is concerned with sparsifying the activations or outputs of the filters. Other approaches such as Dropout \cite{dropout}, Sparseout \cite{sparseout} and Dasnet \cite{dasnet} are techniques where the activations are masked during training to induce activation sparsity. Our sparsity loss does not involve any masking. Methods such as ICNN \cite{icnn}, ICCNN \cite{iccnn} and learning sparse class-specific gate structures \cite{disentangledfilters} deal with class-specific filters but unlike our approach they do not focus on binarization of the filter outputs. Some more approaches induce sparsity in the activations layer-wise i.e. a certain percentage of activations in each layer is retained \cite{activation_sparsity1,activation_sparsity2}. We on the other hand induce class-wise sparsity.

Previous research on interpreting CNNs has primarily focused on visualizing the outputs of CNN layers, aiming to establish relationships between input pixels and neuron outputs. Zeiler et al.~\cite{zeiler2014visualizing} employ output activations as a regularizer to identify discriminative image regions. Other studies~\cite{https://doi.org/10.48550/arxiv.1412.6815,selvaraju2017grad,https://doi.org/10.48550/arxiv.1312.6034}, utilize gradients to perform input-output mapping. Unlike these visualization methods, our approach is useful for methods that generate a rule set using filter outputs. 

\section{Conclusion and Future Work}
In this work, we have addressed the issue of information loss due to the post-binarization of filters in rule-extraction frameworks such as NeSyFOLD. We presented a novel sparsity loss that helps in learning class-specific sparse filters and binarizes the filter outputs during training itself to mitigate post-binarization information loss. We evaluated five training strategies employing the sparsity loss and compared their performance with the baseline, NeSyFOLD-EBP.


As general guidelines for researchers, we recommend using TS2 and TS3 when high fidelity to the original model is required, alongside a balance between accuracy and interpretability. TS4 is best suited for scenarios where minimizing the rule-set size is a priority. Evidently TS2, TS3 and TS4 all outperform the baseline both w.r.t. accuracy and the rule-set size.

Finally, we demonstrated that interpretable neuro-symbolic methods can achieve accuracy levels within $\textbf{3\% - 4\%}$ of the original CNN model, without compromising on interpretability. This establishes these methods as viable and interpretable alternatives to black-box CNN models.

Currently, our approach is tailored to CNN models, and it would be intriguing to investigate whether a similar sparsity loss function could be adapted for Vision Transformers. Although the challenge lies in the absence of filters that directly capture concepts, sparse autoencoders could be used to extract relevant concepts from attention layers \cite{sparse_autoencoders_LLMs}.
Another promising direction is integrating the symbolic rule-set into the training loop to further mitigate information loss by leveraging soft decision tree like structure for gradient backpropogation \cite{soft_decision_tree}. We aim to advance the development of interpretable neuro-symbolic models that match or even surpass the performance of black-box neural models, continuing our quest for models that combine interpretability and accuracy.
\bibliographystyle{named}
\bibliography{ijcai25}
\end{document}